\documentclass[preprints,article,accept,moreauthors,pdftex]{Definitions/mdpi} 

\usepackage{rotating}
\usepackage{subcaption}
\usepackage{lscape}
\firstpage{1} 
\pubvolume{xx}
\issuenum{1}
\articlenumber{5}
\pubyear{2020}
\copyrightyear{2020}
\history{Received: date; Accepted: date; Published: date}

\Title{An Odor Labeling Convolutional Encoder-Decoder for Odor Sensing in Machine Olfaction}

\Author{Tengteng Wen $^{1, *}$\orcidA{}, Zhuofeng Mo $^{1}$, Jingshan Li $^{1}$, Qi Liu $^{1}$, Liming Wu $^{1}$ and Dehan Luo $^{1}$}

\AuthorNames{Tengteng Wen, Zhuofeng Mo, Jingshan Li, Qi Liu, Liming Wu and Dehan Luo}

\address{%
$^{1}$ \quad Department of Electromechanical Engineering, Guangdong University of Technology, 100, Waihuan Rd. W., Guangzhou Higher Education Mega Center, Guangzhou, China}

\corres{Author to whom correspondence should be addressed.}

\abstract{Deep learning methods have been widely applied to visual and acoustic technology. In this paper, we proposed an odor labeling convolutional encoder-decoder (OLCE) for odor identification in machine olfaction. OLCE composes a convolutional neural network encoder and decoder where the encoder output is constrained to odor labels. An electronic nose was used for the data collection of gas responses followed by a normative experimental procedure. Several evaluation indexes were calculated to evaluate the algorithm effectiveness: accuracy $92.57\%$, precision $92.29\%$, recall rate $92.06\%$, F1-Score $91.96\%$, and Kappa coefficient $90.76\%$. We also compared the model with some algorithms used in machine olfaction. The comparison result demonstrated that OLCE had the best performance among these algorithms.}

\keyword{Machine Olfactions; Odor Identifications; Electronic Nose; Neural Networks; Encoder-decoder}  

\begin{document}

\section{Introduction}

Machine olfaction is the advanced technology that captures odorous materials and identifies them by distinguishing the differences in response patterns. Usually electronic noses (e-noses) are used, which consist of an array of gas sensors and intelligent identification algorithms mimicking biological noses, to 'smell' and 'sense' odors \cite{Gardner.1994, Wasilewski.2019}.

Gas sensors detect gases by usually measuring the change in electrical conductivity. Sensitivity, selectivity, response time, and recovery time are the major specifications to evaluate the performance of a gas sensor \cite{Dey.2018}. There are different types of gas sensors: catalytic combustion, electrochemical, thermal-conductive, infrared absorption, paramagnetic, solid electrolyte, and metal oxide semiconductor sensors \cite{Dey.2018}. In recent years, paper-based sensors, which are a new type of gas sensor fabricated by cellulose paper, have the characteristics of flexibility, low-cost, lightweight, tailorablity, and environmental-friendly \cite{Tai.2020}. The response of a gas sensor detecting an odor is a synthetical process since the sensor may be sensitive to a group of different molecules, which is usually called 'cross-sensitivity'. Cross-sensitivity is a characteristic of gas sensors because of poor selectivity \cite{Feng.2019}. It is an interference of measuring gas concentration using a single gas sensor. However, it can be utilized as a feature to identify odors when an array of gas sensors detects odors. Response patterns of sensor signals are different from various odors. It is difficult to interpret sensing responses due to the synthetically non-linear sensing process of gas sensors. Most gas sensors are fabricated for detecting industrial gases or volatile organic chemicals (VOCs).

Odor identifications have been well developed in recent years and have been applied to specific fields. However, such methods ignore the essence of odors. An odor is usually composed of a group of odorous compounds. We human beings sniff the odorous mixture, discriminate, and identify the odor if people are trained to learn the odor. We have difficulties describing an unknown odor without prior knowledge. Instead, we describe it by using some semantic words. Accordingly, is there a method to describe the odor space so that odor can be recorded and encoded in some general forms?

It is a challenge to determine the dimensionality of the olfactory perceptual space because there are still a lot of efforts required in investigating the mechanism of olfactory perceptions. Physiological studies had identified that the human olfactory system consists of around 400 odorant receptor types \cite{Malnic.2004}. An odor activates some of these odorant receptor types to generate a specific pattern so that humans can discriminate against it. The number of odorant receptor types sets the upper bound on the dimensionality of the perceptual space. There are not dedicated vocabulary to describe odors in major languages. Instead, words about objects, for example, flowers and animals, or emotions such as pleasantness are applied to describe olfactory perceptions. J. E. Amoore claimed that odors were divided into 7 groups which were regarded as primary odors \cite{John.1977}. Markus Meister suggested that olfactory perceptual space may contain around 20 dimensions or less \cite{Meister.2015}, and Yaara and Noam reviewed that humans are good at odor detection and discrimination, but are bad at odor identification and naming \cite{Yeshurun.2010}. Semantic descriptors profiled from a list of defined verbal words are rated by human sniffers. Up to now, there is not a universal list of odor semantic descriptors yet.

Currently, there is not an odor space to describe the variety of odors in nature. Some studies revealed a significant relationship between odor molecular structure information and olfactory perceptions \cite{Karen.1996}. Functional groups and hydrocarbon structural features were considered to be factors influencing olfactory perceptions. A hypothesis demonstrated that odorants possessing the same functional groups activate the same glomerular modules \cite{Johnson.2002} which generate similar perceptual patterns so that humans identify them as the same type of odor. Recent studies revealed that 3D structure information of odorous molecules has a more noticeable impact on olfactory perceptions \cite{Rojas.2015}. Considering the complexity of molecular structure information, the mapping to odor space may be non-linear.

Several studies investigated the map between odor responses and odorous perceptual labels. T. Nakamoto designed an odor sensing system that consisted of a mass spectrum and large-scale neural networks to predict odor perceptual information \cite{Nakamoto.2019}. R. Haddad et al. investigated the relationship between odor pleasantness and e-nose sensing responses by modeling a feed-forward back-propagation neural network \cite{Haddad.2010}. D. Wu et al. designed a convolutional neural network for predicting odor pleasantness \cite{Wu.2019}. These models used in predicting odor perceptual descriptors perform decently in some particular datasets. However, machine percepts and describes odors using distributed representation is still a challenge for us.

It is worth establishing some forms of odor space to describe a sufficiently complete group of odors in nature. An odor space should be some form of numerical values with definite dimensionality. The odor space should be a linear space for convenient interpreting because of the non-linear map. Those semantic olfactory descriptors are only some points in the quantization just as the color "red" is quantified to (255, 0, 0) in RGB color space. The importance of such odor space is a quantization form so that odors can be converted to information for data storage or transmission. The odors can be reproduced by blending some similar odorants to generate the odor.

Machine olfactions have been applied widely to many fields in recent years. Some linear methods such as principal component analysis (PCA), linear discriminant analysis (LDA), support vector machines (SVM), etc. were used in the analysis of odor discrimination \cite{Marco.2012}. PCA is an unsupervised method ignoring discriminant information which is a popular method for dimensionality reduction \cite{Bedoui.2013}. LDA is a supervised method for classification by finding decision surfaces and calculating the signed orthogonal distance of data points. It has been used in the identification of Chinese herbal medicines \cite{Luo.2012}. SVM is some kind of regularization in which the aim is to find the maximum margin between classes. K. Brudzewski applied SVM as the classification tool for identifying tobacco \cite{Brudzewski.2012}. Classifiers using linear methods can be transferred to convex problems which have the advantages of mathematical interpretability. Non-linear methods such as artificial neural networks (ANN) were also introduced in machine olfactions. In recent years, deep learning methods are dramatically developing and widely used in various fields such as computer visions, speech processing, automatic driving, etc. They also have been introduced in machine olfaction for odor identifications \cite{Wu.2019,Jong.2019}.

There are several advantages that machine olfaction technology applies to many fields. Firstly, it is a non-destructive technique to detect volatiles released from the surface of objects \cite{Brezmes.2005b}. Secondly, e-nose is usually portable which is convenient to detect odors anywhere and anytime \cite{Das.2009}. Thirdly, e-noses have the capacity of extending out olfactory perception scopes since gas sensors are capable of detecting those chemicals which humans are unable to smell and sense \cite{Deshmukh.2015}. Furthermore, e-noses can be used in some unpleasant environments \cite{Murphy.2014,Li.2019}.

Linear methods for classifications usually require highly correlated features and high calibration costs which the number of training data is limited \cite{Marco.2012}. Non-linear methods have difficulties in interpretation. Nonetheless, non-linear methods especially deep learning methods have a higher capacity of identifying more odors. In this paper, we borrowed the idea from auto-encoder and proposed a novel deep learning algorithm for odor identification - Odor Labeling Convolutional Encoder-Decoder (OLCE). OLCE consists of an encoder and a decoder, where the encoder output is constrained to odor labels. OLCE has a decoder structure which offers some clues on how the model learns features. In the following paragraphs, we will first describe the experimental setups and the modeling of OLCE. After that, the performance of the model, comparison with other methods, and an overview of decoded response results will be illustrated. Furthermore, the perspective of machine olfaction will be discussed.

\section{Materials and Methods}

\subsection{Research scheme}
\begin{figure}[H]
  \centering
  \includegraphics[width=\linewidth]{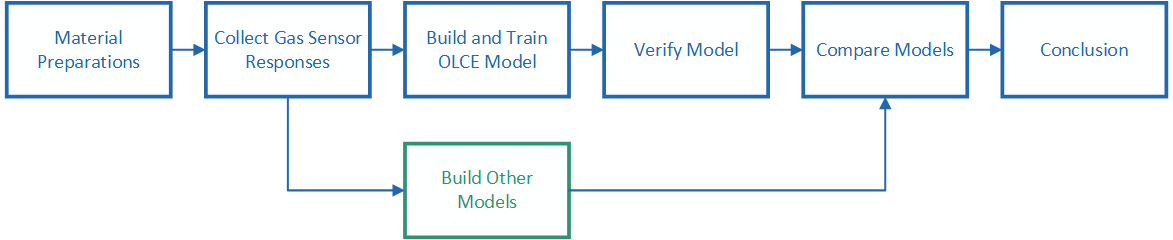}
  \caption{OLCE workflow.}
  \label{fig_scheme}
\end{figure}

The OLCE model was built, trained, and tested by self-collecting gas response datasets. In the study, odors from seven non-crushed Chinese herbal medicines were collected by an electronic nose. We kept experimental settings of gas-response collections by a self-designed standard procedure to control the detecting consistency and the data effectiveness. We built other algorithms that had already been used for odor identification to examine the performance of OLCE. The procedure of the study was displayed in Figure \ref{fig_scheme}.

\subsection{Experiment Setup}
\begin{figure}[H]
  \centering
  \includegraphics[width=\textwidth]{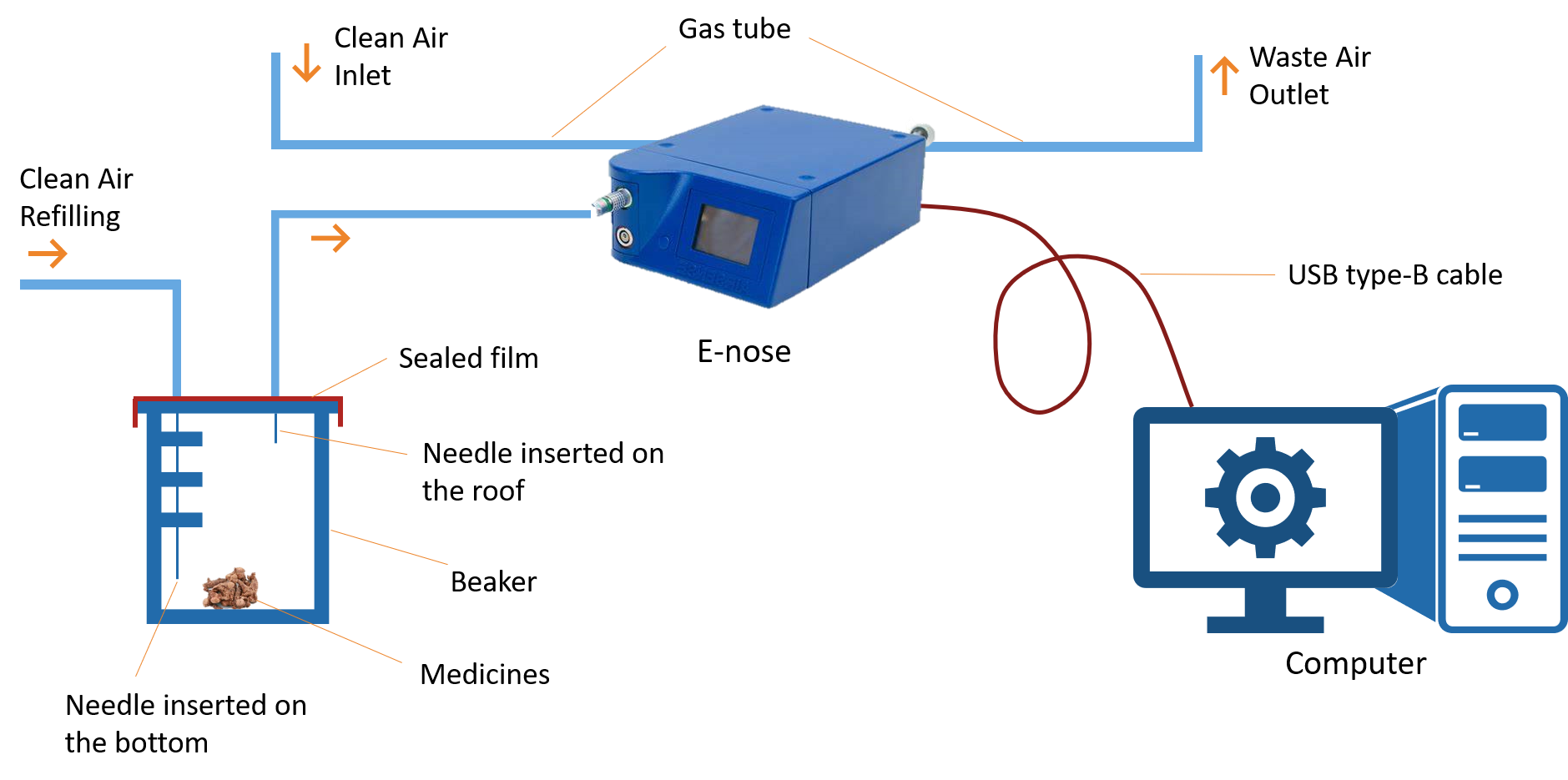}
  \caption{The architecture of the experimental setup for collecting odor response data. Chinese Herbal medicines were selected as the experimental materials, and they were placed in clean beakers covered by a sealed film. A needle was inserted to the bottom for clean-air refilling. Another needle was inserted on the roof of the beaker beneath the sealed film for collecting headspace gases. The clean-air inlet was connected to the purge gas port on the PEN-3 e-nose for flushing gas sensors, and the waste-air outlet was connected to the waste port for ejecting waste gases. Response data were collected by e-nose and transmitted to a computer via a USB Type-B cable connecting e-nose and computer.}
  \label{fig_exp}
\end{figure}

The instrument and tools used in the experiment included a PEN-3 electronic nose, beakers, and a computer. Experimental subjects were placed in beakers for the data collection. The PEN-3 electronic nose manufactured by AirSense Inc. was used for collecting gas sensor responses. The computer with installed Winmuster, which is the PEN-3 e-nose control software designed by AirSense Inc., was used to connect and control the e-nose. The architecture experimental setup is displayed in Figure \ref{fig_exp}.

\subsection{The Preparation of Experimental Materials}
We selected seven Chinese herbal medicines (Betel Nut, Dried Ginger, Rhizoma Alpiniae Officinarum, Tree Peony Bark, Fructus Amomi, Rhioxma Curcumae Aeruginosae, Fructus Aurantii) for the experiment. To ensure the consistency of gas sensor responses, the procedures for preparing these materials were carefully set as follows:

\begin{enumerate}
  \item Materials in initial conditions are placed in clean beakers separately.
  \item Beakers are equilibrated for over 20 minutes {color{red}for enrichment} of volatiles released from the surface of medicines.
  \item The temperature is kept around 25 $^{\circ}$C.
  \item The humidity is kept around 75\%.
\end{enumerate}

\subsection{PEN-3 Electronic Nose}
Response data were collected from an e-nose, PEN-3, AirSense Inc. The PEN-3 e-nose is a general-purpose gas response signal sampling instrument with 10 metal-oxide gas sensors each of which has different sensitivity to a different gases as shown in Table \ref{tab_pen3}. Since the combination of these 10 sensors, PEN-3 has the ability to sense various gases, which makes it a suitable instrument for the research. The settings of the e-nose are illustrated in Table \ref{tab_settings}.

\begin{table}[H]
  \caption{Descriptions of the sensor array in PEN-3 e-nose.}
  \centering
  \begin{tabular}{lp{13.7cm}}
  \toprule
  \textbf{Sensor} & \textbf{Sensor Sensitivity and General Description}	\\
  \midrule
  W1C & Aromatic compounds. \\
  W5S & Very sensitive, broad range of sensitivity, reacts to nitrogen oxides, very sensitive with negative signals. \\
  W3C & Ammonia, used as sensor for aromatic compounds. \\
  W6S & Mainly hydrogen. \\
  W5C & Alkanes, aromatic compounds, less polar compounds. \\
  W1S & Sensitive to methane. Broad range. \\
  W1W & Reacts to sulphur compounds, H2S. Otherwise sensitive to many terpenes and sulphur-containing organic compounds. \\
  W2S & Detects alcohol, partially aromatic compounds, broad range. \\
  W2W & Aromatic compounds, sulphur organic compounds. \\
  W3S & Reacts to high concentrations ($>$100 mg/kg) of methane--aliphatic compounds. \\
  \bottomrule
  \end{tabular}
  \label{tab_pen3}
\end{table}

\begin{table}[H]
  \caption{Settings of PEN-3 electronic nose.}
  \centering
  \begin{tabular}{rl}
  \toprule
  \textbf{Options} & \textbf{Settings}	\\
  \midrule
  Sample interval         & 1.0 sec \\
  Presampling time        & 5.0 sec \\
  Zero point trim time    & 5.0 sec \\
  Measurement time        & 120 sec \\
  Flushing time           & 120 sec \\
  Chamber flow            & 150 ml/min \\
  Initial injection flow  & 150 ml/min \\
  \bottomrule
  \end{tabular}
  \label{tab_settings}
\end{table}

\subsection{OLCE Modeling}

\begin{figure}[H]
  \centering
  \includegraphics[width=\linewidth]{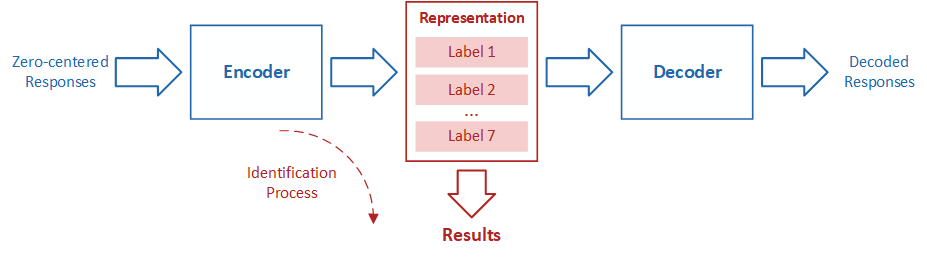}
  \caption{The basic principle of OLCE.}
  \label{fig_olce}
\end{figure}

Figure \ref{fig_olce} describes the principle of OLCE. OLCE contains a convolutional encoder and a convolutional decoder. The OLCE input is those responses that have been zero-center normalized. The OLCE output aims to reproduce the input. The intermediate layer is a representation that outputs the identification results. The encoder and the decoder are trained together using a training dataset. To verify the model, the results in the representation layer are used to evaluate the performance of the model.

The original response data were firstly zero-center normalized, then sent into the OLCE model. The $i$-th zero-center normalized response data point $x'_i$ is calculated as follow:

\begin{equation}
  x'_i = \frac{x_i-x_{mean}}{x_{max}-x_{min}},
\end{equation}
where $x_i$ is the $i$-th original data from e-nose, $x_{mean}$ is the average value of 120 data points collected from a gas sensor, and $x_{max}$ and $x_{min}$ are the maximum and the minimum value of the 120 data points, respectively. 

Suppose the input is $\mathbf{X}$ which here is the sensing response collected by gas sensors. The labels of Chinese herbal medicines are defined as $\mathbf{y}$ which is one-hot encoding. The encoder is defined as $\mathcal{F}(\bullet)$ and the decoder is defined as $\mathcal{G}(\bullet)$. So the encoder can be presented as

\begin{equation}
  \mathbf{y} = \mathcal{F}(\mathbf{X}),
\end{equation}

and the decoder can be presented as

\begin{equation}
  \mathbf{X}' = \mathcal{G}(\mathbf{y}),
\end{equation}

where $\mathbf{X}'$ output of the decoder. The aim of building the encoder-decoder is to gain an accurate labeling results $\mathbf{y}$. To achieve this, the re-build responses $\mathbf{X}'$ must approximate to the original responses $\mathbf{X}$: $\mathbf{X}' \rightarrow \mathbf{X}$. In other words, the aim of the encoder-decoder can be illustrated as follow:

\begin{equation}
  \begin{aligned}
    \mathnormal{minimize} \quad & y-\mathcal{Y} \\
    \mathnormal{subject \, to} \quad & \mathcal{G}(\mathcal{F}(\mathbf{X}))-\mathbf{X}.
  \end{aligned}
\end{equation}

The encoder was designed with a convolutional neural network. The convolutional layer extract features by computing the product sum of the input variables. $ReLU$ was used to introduce non-linearity in the convolutional network.
\begin{equation}
  ReLU(x) = max(0, x).
\end{equation}
A max-pooling layer was introduced to reduce spatial size of the convolved data. After that, a fully connected layer was introduced to learn non-linear combinations of the high-level features. Softmax was implemented through the output layer as a classifier to identify odor labels. Symmetrically, the decoder was a convolutional neural network with the same structure. Figure \ref{fig_olce_struct} describes the architecture of the encoder and decoder, while Table \ref{tab_olce-par} shows the network parameters.

\begin{landscape}
\begin{figure}[H]
  \centering
  \includegraphics[width=0.8\linewidth]{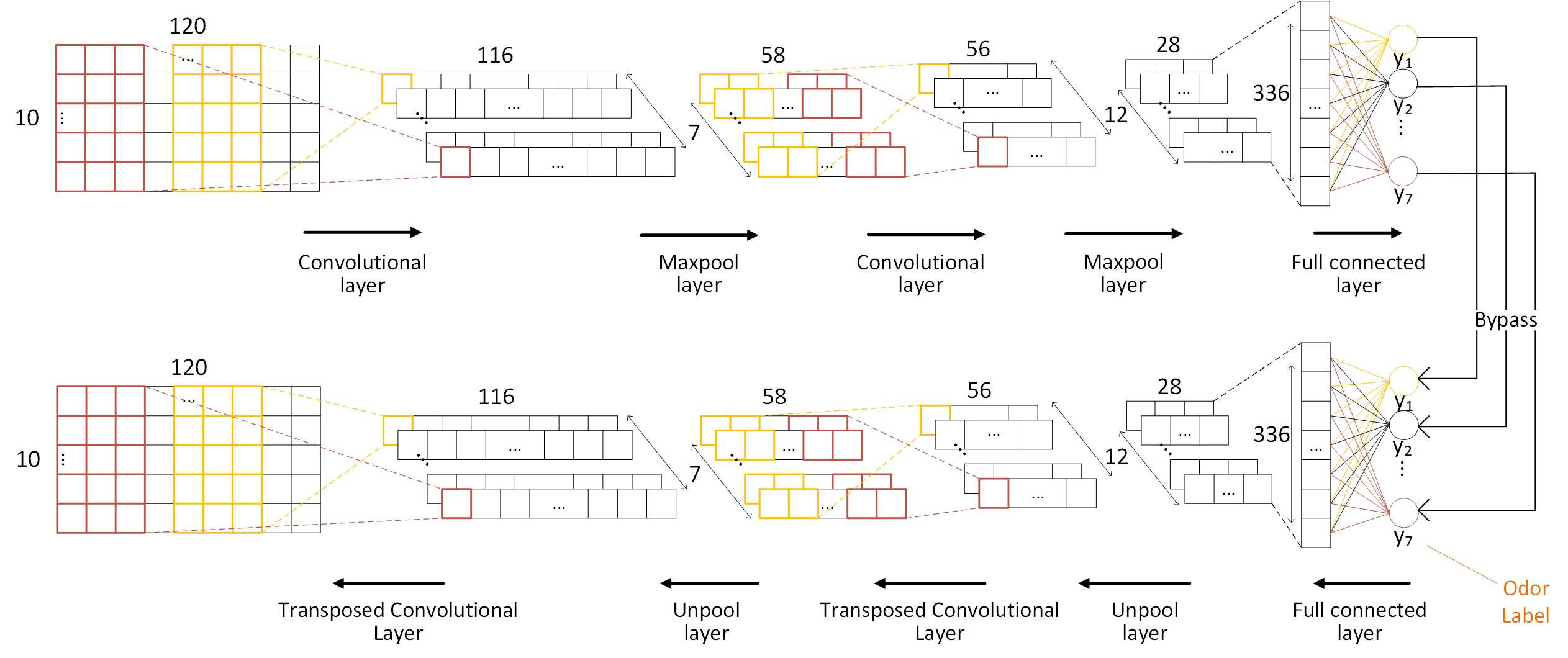}
  \caption{Architecture of the odor labeling convolutional encoder-decoder.}
  \label{fig_olce_struct}
\end{figure}

\begin{table}[H]
  \caption{Structural parameters of the odor labeling convolutional encoder-decoder.}
  \centering
  \begin{tabular}{cccc}
  \toprule
  \textbf{Layer} & \textbf{Type} & \textbf{Filter Shape} & \textbf{Input Size}	\\
  \midrule
  Conv1            & conv    & 7*1*5  & 10*1*120 \\
	                 & Maxpool & 1*2    & 7*1*116 \\
  Conv2            & conv    & 12*1*3 & 7*1*58 \\
	                 & Maxpool & 1*2    & 12*1*56 \\
  FC3              & FC      & 7*336  & 12*1*28 \\
  Classifier       & Softmax & -      & 7 \\
  FC3              & FC      & 336*7  & 7 \\
	                 & Unpool  & 1*2    & 12*1*28 \\
  Transposed Conv2 & Transposed conv2 & 7*1*3  & 12*1*56 \\
	                 & Unpool           & 1*2    & 7*1*58 \\
  Transposed Conv1 & Transposed conv1 & 10*1*5 & 7*1*116 \\
  \bottomrule
  \end{tabular}
  \label{tab_olce-par}
\end{table}
\end{landscape}

\subsection{Comparison Models}

In order to take a view on the performance of OLCE, several algorithms that had been applied to machine olfactions were selected for comparison.
\begin{itemize}
	\item linear discriminant analysis (LDA) \cite{Akbar.2016},
	\item multi-layer perception (MLP) \cite{Benrekia.2013},
	\item decision tree (DT) \cite{AitSiAli.2017},
	\item principle component analysis (PCA) with LDA \cite{Sun.2018},
	\item convolutional neural networks (CNN) and support vector machine (SVM) \cite{YanShi.2019}.
\end{itemize}

LDA can be used not only for dimensionality reduction but also for classification. LDA reduces in-class distances and increases the distances between classes.

MLP classifier is an artificial neural network and has been applied to odor identification. MLP is is a supervised non-linear function approximator learns a function $f(\bullet): R^{m}\rightarrow R^{n}$, where $m=1200$ was a $120*10$ sample and $n=7$ is the labels. The MLP consisted of 4 hidden layers with the ReLU activation function.

DT is a non-parametric supervised learner which classifies data based on already-known sample distribution probability. It performed decently in odor classification. We here set the classification criterion to Gini,
\begin{equation}
  H(X_m)=\sum_kp_{mk}(1-p_{mk}),
\end{equation}
where $X_m$ is samples used the node $m$. The proportion of class $k$ in node $m$ is $p_{mk} = 1/N \sum_{x_i\in R}I(y_i=k)$. It represents a region $R$ with $N$ observations. In order to prevent overfitting, the maximum depth of a tree was limited to 10.

PCA-LDA is a combination model and has been applied to odor identification. PCA implemented the dimensionality reduction through orthogonally projecting input data onto a lower-dimensional linear space by singular value decomposition with scaling each component. LDA was implemented for the classification.

In the CNN-SVM model, CNN is a typical feed-forward neural network for feature extraction. SVM is a supervised learning algorithm for classification. The CNN consisted of 2 one-dimension convolutional layers, fully max-pooling layers, and a fully-connected layer.

All models were coded in Python and open-source packages scikit-learn \cite{scikit-learn} and PyTorch \cite{NEURIPS2019_9015} were used to build models.

\section{Results}

\subsection{The Input of OLCE}

The input of OLCE is a zero-centered $10\times 120$ dataset which is collected by PEN-3 e-nose. The gas sensor responses were collected by PEN-3 e-nose followed by the experimental procedure illustrated in the previous section. For each medicine, 100 response samples were measured so that the total number of samples in the dataset is $7\times100=700$. Each sample is actually a $120\times10$ matrix.

\begin{landscape}
\begin{figure}[htbp]
  \centering
  \includegraphics[width=\linewidth]{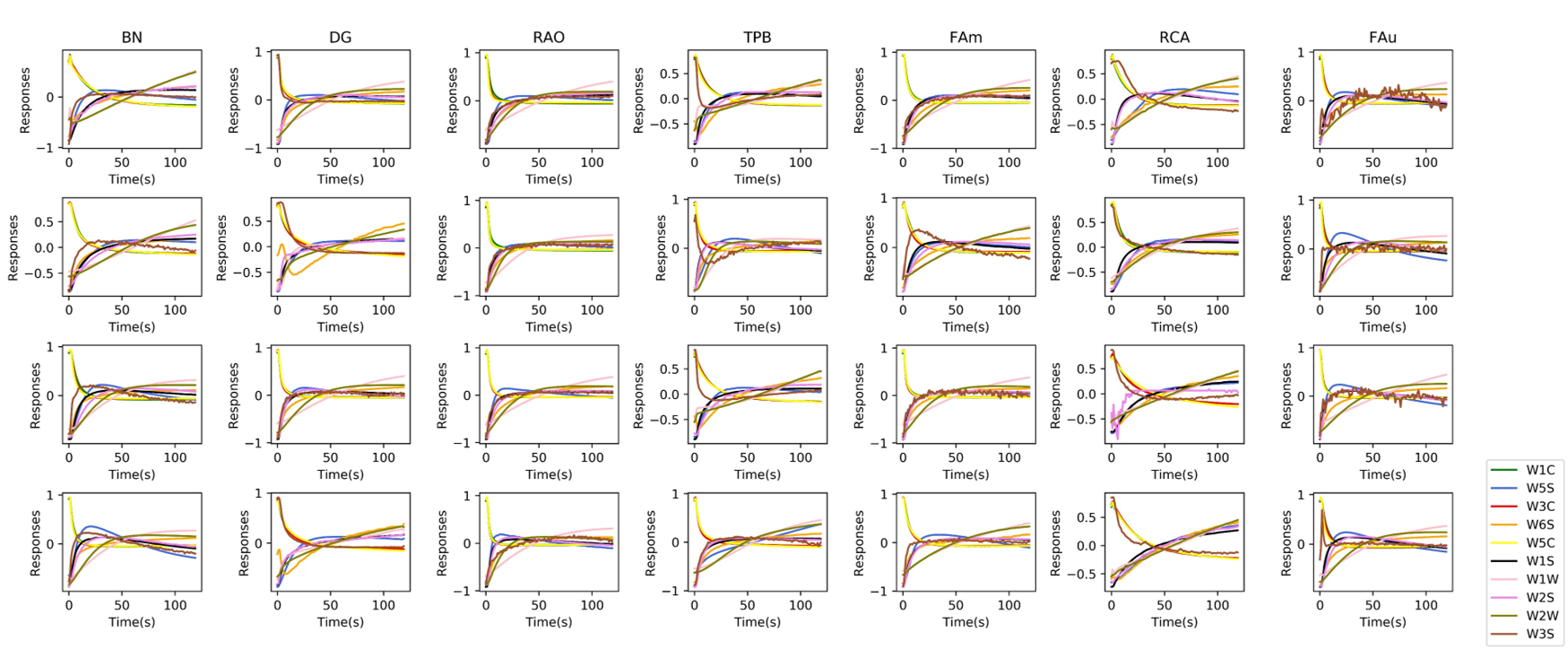}
  \caption{Centralized sensing responses of seven types of Chinese herbal medicines (Betel Nut (BN), Dried Ginger (DG), Rhizoma Alpiniae Officinarum (RAO), Tree Peony Bark (TPB), Fructus Amomi (FAm), Rhioxma Curcumae Aeruginosae (RCA), Fructus Aurantii (FAu)). We randomly selected 4 response samples of each medicine. All response data were implemented by centralized normalization.}
  \label{fig_responses}
\end{figure}
\end{landscape}

Figure \ref{fig_responses} compares the zero-centered responses to the input of OLCE. We randomly selected 4 samples from each medicine class. It can be seen that there are some slight differences in the same medicine class because of different within-class medicine sources used for the collection experiment. Responses have noticeable differences between classes. Some sensors show upwards baseline drift because various volatilization rates of some volatiles and their sensitivities to volatiles. Some sensors show downwards baseline drift because of the overflow in the sensor chamber. Since an OLCE receives a $10\times 120$ sample as an input dataset without feature extraction, these drifts can be ignored.

\subsection{OLCE Evaluation}
Each OLCE model was executed 10 times and several performance evaluation indexes (accuracy, precision, recall rate, F-score, Kappa rate, and Hamming loss) were calculated to view the model effectiveness. The results were displayed in Table \ref{tab_olce-results}.

OLCE had the maximum accuracy $0.96$ and minimum accuracy $0.8457$. It had a decent precision rate (between $0.8397$ and $0.9635$) and recall rate ($0.8419$ and $0.9576$). The $F_1$ score of the model was between $0.8379$ and $0.9591$. It demonstrated that OLCE had less false positive and false negative predicting output.

Kappa coefficient was also calculated to evaluate the consistency and classifier precision.
\begin{equation}
  \textnormal{Kappa} = \frac{P_o - P_e}{1 - P_e},
\end{equation}
where $P_o$ is the accuracy and $P_e$ is calculated as follow:
\begin{equation}
  P_e = \frac{a_1*b_1+a_2*b_2+...+a_7*b_7}{n*n},
\end{equation}
where $i=1, 2, ..., 7$ is the class index, $a_i$ represents the accumulated amount of samples of each class in the dataset, $b_i$ represents the accumulated number of samples in each class after the classification, and $n$ is the total number of samples. The Kappa revealed that OLCE has excellent consistency.

\begin{table}[htbp]
  \centering
  \caption{Performance evaluation indexes for OLCE model.}
    \begin{tabular}{cccccc}
    \toprule
    No.   & Accuracy & Precision & Recall & F1 Score & Kappa \\
    \midrule
    1     & 0.9142  & 0.9269  & 0.9276  & 0.9249  & 0.9130  \\
    2     & 0.8800  & 0.9635  & 0.9576  & 0.9584  & 0.9533  \\
    3     & 0.9485  & 0.8858  & 0.8624  & 0.8691  & 0.8520  \\
    4     & 0.9428  & 0.9312  & 0.9347  & 0.9320  & 0.9197  \\
    5     & 0.9714  & 0.9163  & 0.9157  & 0.9129  & 0.8998  \\
    6     & 0.9200  & 0.9333  & 0.9354  & 0.9330  & 0.9196  \\
    7     & 0.8971  & 0.9599  & 0.9590  & 0.9591  & 0.9532  \\
    8     & 0.9428  & 0.8397  & 0.8419  & 0.8379  & 0.8193  \\
    9     & 0.9485  & 0.9404  & 0.9404  & 0.9395  & 0.9264  \\
    10    & 0.8914  & 0.9317  & 0.9314  & 0.9296  & 0.9198  \\
    \midrule
    Average & 0.9257  & 0.9229  & 0.9206  & 0.9196  & 0.9076  \\
    \bottomrule
    \end{tabular}%
  \label{tab_olce-results}%
\end{table}%

\subsection{Comparison}

Several algorithms used in machine olfactions were built to compare OLCE. Each model was executed 10 times, and the accuracy scores of each model are illustrated in Table \ref{tab_comp-results}. It can be seen that the highest and lowest accuracy of LDA is $0.9314$ and $0.8686$. CNN-SVM has the highest and lowest scores of $0.9371$ and $0.8514$. MLP and PCA-LDA has relatively lower scores which the best scores were $0.4342$ and $0.5200$, respectively. The decision tree yielded relatively good scores between $0.7600$ and $0.8857$. It can be seen from the 'Max.' and 'Min.' columns that OLCE has the best scores (the highest score is $0.9714$ and the lowest one is $0.8800$. Moreover, OLCE has the best average score ($0.9257$). The 'Var.' column describes the variances of the accuracy scores from the 10 models of each algorithm. It can be seen that LDA has the highest consistency because of the lowest accuracy variance ($0.0005$) between 10 LDA models. On contrary, the PCA-LDA model got the highest variance of $0.0141$ which reveals the worst training consistency. It can be noted that OLCE has the third-lowest accuracy variance that is $0.0009$.

Overall, as the results illustrate above, LDA, CNN-SVM, and OLCE has a decent performance for machine olfaction according to better average predicting accuracy and stable consistency. Moreover, considering the comprehensive advantages in accuracy, precision, recall rate, $F_1$ score, and Kappa coefficients, OLCE is sufficiently suitable to discriminate odors from gas responses collected by e-nose.

\subsection{Overview of decoded responses}
OLCE is an encoder-decoder structure model, and the representation layer consists of several odor labels. It is interesting to take a view on decoded responses. We randomly selected one original response and one decoded response from both the training set and test set.

\begin{figure}[htbp]
  \centering
  \begin{subfigure}[b]{0.49\linewidth}
      \includegraphics[width=\textwidth]{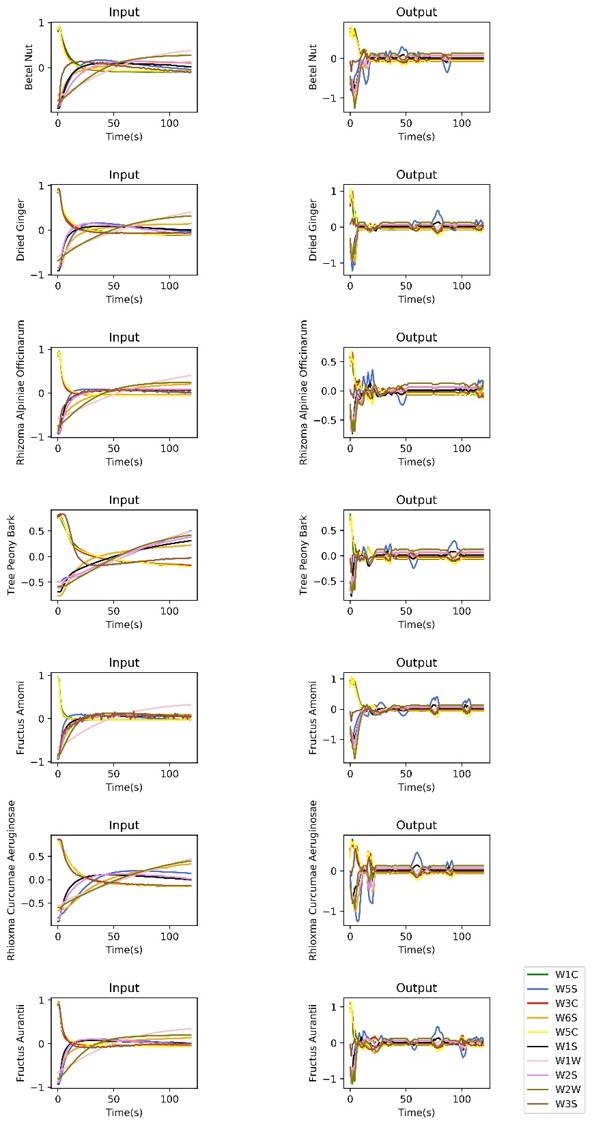}
      \caption{Training Set}
      \label{fig_dec-res_train}
  \end{subfigure}
  ~ 
  \begin{subfigure}[b]{0.49\linewidth}
      \includegraphics[width=\textwidth]{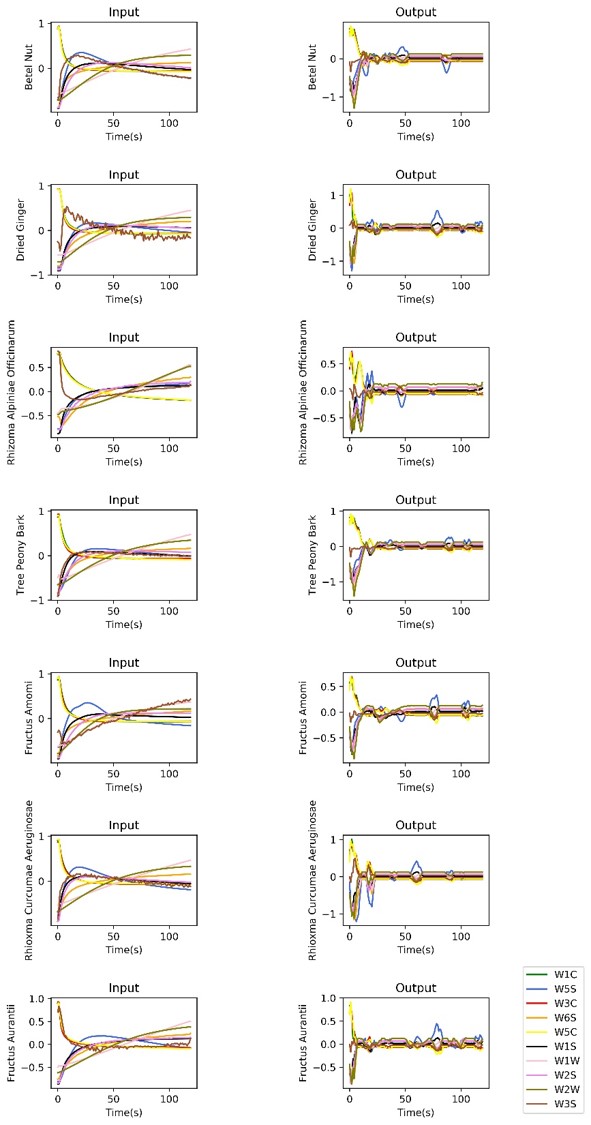}
      \caption{Test set}
      \label{fig_dec-res_test}
  \end{subfigure}

  \bigskip
  \begin{subfigure}[b]{0.8\textwidth}
    \includegraphics[width=\textwidth]{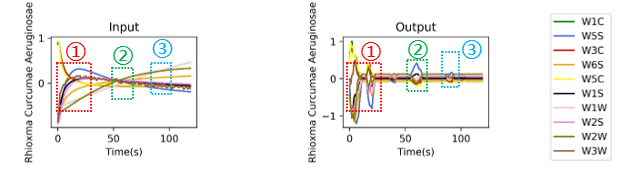}
    \caption{A typical response}
    \label{fig_dec-res_analyze}
\end{subfigure}
  \caption{The original responses and decoded responses. In \ref{fig_dec-res_train} and \ref{fig_dec-res_test}, the left column shows the input responses of OLCE encoder, and the right column describes the output responses of OLCE decoder. The subfigure \ref{fig_dec-res_analyze} highlights the significant features extracted by decoder.}
  \label{fig_dec_responses}
\end{figure}

Figure \ref{fig_dec-res_train} and \ref{fig_dec-res_test} shows the comparison of encoder input and decoder output. Firstly, OLCE reproduces response signals in response state. Some gentle response changes in the response states are decoded as some fluctuating signals. For instance, in Figure \ref{fig_dec-res_train}, row 4, some baseline drifts are decoded as some fluctuating signals. Secondly, OLCE focuses on positive or negative baseline drift. For example, in Figure \ref{fig_dec-res_test}, row 5, when the signal changes accumulate exceed a certain level, the decoder generates some fluctuations. It is possible that some response changes may activate OLCE to generate fluctuating waves. These fluctuations can be regarded as 'feature stamps'. These feature stamps reveal some clues of which features OLCE focuses on. Furthermore, the model ignores response fluctuations from one single gas sensor. For instance, as shown in Figure \ref{fig_dec-res_train}, the "W3S" response (brown line) in subfigure "Fructus Amomi" (row 5, column 1) fluctuates obviously, but the decoder did not take it as a feature.

Figure \ref{fig_dec-res_analyze} describes a typical decoded response. It can be seen that OLCE learns features using one or more small windows in a response dataset. The response state is the most significant feature for OLCE, as shown the red dotted box. Moreover, the OLCE regards some gentle changes in steady state as some features. The intersections of curves can be also some significant features as shown the green dotted box. OLCE may also concentrate on those accumulated changes in steady state as shown the blue dotted box. 

\begin{landscape}
  \begin{table}[p]
    \centering
    \caption{Accuracy scores of 6 models (LDA, MLP, DT, PCA+LDA, CNN+SVM, OLCE). In PCA-LDA model, grid search for finding the best number of dimensions using PCA, which reduced to 49 dimensions.}
    \begin{tabular}{ccccccccccccccc}
    \toprule
    \multirow{2}[4]{*}{Models} & \multicolumn{10}{c}{Predictions}                                              & \multirow{2}[4]{*}{Max.} & \multirow{2}[4]{*}{Min.} & \multirow{2}[4]{*}{Ave.} & \multirow{2}[4]{*}{Var.} \\
  \cmidrule{2-11}          & 1st   & 2nd   & 3rd   & 4th   & 5th   & 6th   & 7th   & 8th   & 9th   & 10th  &       &       &       &  \\
    \midrule
    LDA   & 0.9029  & 0.9257  & 0.9314  & 0.8686  & 0.8971  & 0.9029  & \textcolor[rgb]{ 1,  0,  0}{0.9314 } & 0.9200  & 0.9086  & 0.8800  & 0.9314  & 0.8686  & 0.9069  & \textcolor[rgb]{ 1,  0,  0}{0.0005 } \\
    MLP   & \textcolor[rgb]{ 1,  0,  0}{0.4342 } & 0.2114  & 0.1200  & 0.1542  & 0.3771  & 0.2628  & 0.2285  & 0.1428  & 0.1542  & 0.2800  & 0.4342  & 0.1200  & 0.2365  & 0.0109  \\
    DT    & 0.8629  & 0.8114  & 0.8514  & 0.7600  & 0.8514  & 0.7943  & 0.8400  & 0.8229  & \textcolor[rgb]{ 1,  0,  0}{0.8857 } & 0.8171  & 0.8857  & 0.7600  & 0.8297  & 0.0013  \\
    PCA-LDA & 0.2857  & 0.4342  & 0.3200  & \textcolor[rgb]{ 1,  0,  0}{0.5200 } & 0.4057  & 0.1542  & 0.4400  & 0.1828  & 0.4342  & 0.3200  & 0.5200  & 0.1542  & 0.3497  & 0.0141  \\
    CNN-SVM & \textcolor[rgb]{ 1,  0,  0}{0.9371 } & 0.9085  & 0.9142  & 0.9028  & 0.9314  & 0.9085  & 0.9028  & 0.8514  & 0.9314  & 0.9085  & 0.9371  & 0.8514  & 0.9097  & 0.0006  \\
    OLCE  & 0.9142  & 0.8800  & 0.9485  & 0.9428  & \textcolor[rgb]{ 1,  0,  0}{0.9714 } & 0.9200  & 0.8971  & 0.9428  & 0.9485  & 0.8914  & \textcolor[rgb]{ 1,  0,  0}{0.9714 } & \textcolor[rgb]{ 1,  0,  0}{0.8800 } & \textcolor[rgb]{ 1,  0,  0}{0.9257 } & 0.0009  \\
    \bottomrule
    \end{tabular}%
    \label{tab_comp-results}%
  \end{table}
  \end{landscape}

\section{Discussion}

Using an e-nose to identify odor is a process of detecting and discriminating those ingredients that gas sensors are sensitive to. It is different from other measuring instruments such as GC-MS that have the capacity of identifying ingredients of an odor. E-nose with an array of gas sensors and a suitable identification algorithm mimics human olfaction to identify odors which can be applied to many fields where requires fast detections because it has advantages of portability, easy-to-design, and low-cost. Hence a reliable algorithm to discriminate various response patterns is necessary.

OLCE has an elegant and symmetrical structure using a convolutional neural network, which is easy to build the model. The experimental results show that OLCE performs decently in odor identification of Chinese herbal medicines according to several performance indexes. It may also suggest that OLCE can be used in other odor identifications. The OLCE encoder encodes sensor responses to odor labels using a convolutional neural network. The OLCE decoder reproduces sensor responses using a convolutional neural network with a symmetrical structure. The reproduced responses on the decoder side reveal some clues on which features OLCE focuses on. The one-hot encoding labels in the representation layer, the intermediate layer, make the classification more robust than categorical encoding because of the mutual exclusivity of the encoding bits.

OLCE is a multi-class classifier that uses one-hot encoding codes to output the identification results. Multi-class classifiers are suitable to be used in the scenario where the identification category is mutually exclusive. The other type is the multi-label classifier that an instance may belong to more than one class. It is interesting to consider that the one-hot encoding labels in the representation layer of OLCE can be replaced by binary encoding labels so that the model can be used as a multi-label classifier.

\section{Conclusion}

In this paper, we proposed a novel Odor Labeling Convolutional Encoder-decoder (OLCE) for odor identification. OLCE is an encoder-decoder structure using convolutional neural network where the representation layer, the intermediate layer, is constrained to odor labels. To evaluate the effectiveness of the model, several performance evaluation indexes (accuracy, precision, recall rate, F1-score, and Kappa coefficient) were calculated. We also built some common algorithms used in odor identifications to compare the performance. Results demonstrated that OLCE had a decent performance according to the performance evaluation indexes. OLCE has the highest average accuracy score ($0.9257$) and better consistency in training models among these algorithms.

\vspace{6pt} 



\authorcontributions{Conceptualization, Tengteng Wen and Dehan Luo; methodology, Tengteng Wen; software, Jingshan Li; validation, Tengteng Wen, Zhuofeng Mo, and Qi Liu; formal analysis, Tengteng Wen; investigation, Zhuofeng Mo; resources, Jingshan Li; data curation, Qi Liu; writing--original draft preparation, Tengteng Wen; writing--review and editing, Tengteng Wen; visualization, Jingshan Li; supervision, Dehan Luo; project administration, Liming Wu; funding acquisition, Liming Wu. All authors have read and agreed to the published version of the manuscript.}

\funding{The work was funded by National Natural Science Foundation of China grant number 61705045; National Natural Science Foundation of China grant number 61571140; Guangdong Science and Technology Department grant number~2019B101001017.}

\conflictsofinterest{The authors declare no conflict of interest.} 

\abbreviations{The following abbreviations are used in this manuscript:\\

\noindent 
\begin{tabular}{@{}ll}
OLCE & Odor Labeling Convolutional Encoder-decoder \\
LDA  & Linear Discriminant Analysis \\
MLP  & Multi-Layer Perception \\
DT   & Decision Tree \\
PCA  & Principle Component Analysis \\
CNN  & Convolutional Neural Networks \\
SVM  & Support Vector Machine \\
BN   & Betel Nut \\
DG   & Dried Ginger \\
RAO  & Rhizoma Alpiniae Officinarum \\
TPB  & Tree Peony Bark \\
FAm  & Fructus Amomi \\
RCA  & Rhioxma Curcumae Aeruginosae \\
FAu  & Fructus Aurantii
\end{tabular}}



\reftitle{References}


\externalbibliography{yes}
\bibliography{references}





\end{document}